\documentclass{elsarticle}
\usepackage{hyperref}
\usepackage{graphicx}
\usepackage{amsmath, amssymb, amsthm}
\usepackage{bm}
\usepackage[algo2e,algoruled]{algorithm2e}
\usepackage{xcolor}
\usepackage[margin=1.4in]{geometry}
\newcommand{\rev}[1]{{\color{black}#1}}
\let\today\relax 

\begin{document}

\begin{frontmatter}
    \journal{Statistics and Computing}
    \title{Random Forest Kernel for High-Dimension Low Sample Size Classification}
    \author[1,2]{Lucca Portes Cavalheiro}
    \author[1]{Simon bernard}
    \author[2]{Jean Paul Barddal}
    \author[1]{Laurent Heutte}
    \address[1]{Univ Rouen Normandie, LITIS UR 4108, F-76000 Rouen, France}
    \address[2]{Graduate Program in Informatics (PPGIa), Pontif\'{i}cia Universidade Cat\'{o}lica do Paran\'{a} (PUCPR), Curitiba, Brazil}
    \begin{abstract}
        High dimension, low sample size (HDLSS) problems are numerous among real-world applications of machine learning. From medical images to text processing, traditional machine learning algorithms are usually unsuccessful in learning the best possible concept from such data. In a previous work, we proposed a dissimilarity-based approach for multi-view classification, the Random Forest Dissimilarity (RFD), that perfoms state-of-the-art results for such problems. In this work, we transpose the core principle of this approach to solving HDLSS classification problems, by using the RF similarity measure as a learned precomputed SVM kernel (RFSVM). We show that such a learned similarity measure is particularly suited and accurate for this classification context. Experiments conducted on 40 public HDLSS classification datasets, supported by rigorous statistical analyses, show that the RFSVM method outperforms existing methods for the majority of HDLSS problems and remains at the same time very competitive for low or non-HDLSS problems.
    \end{abstract}
\end{frontmatter}


\section{Introduction} 

In many modern machine learning problems, data are made available as small samples while being described in high dimensions. These datasets are generally referred to as ``High Dimension, Low Sample Size'' (HDLSS) datasets. These situations occur when the data are intrinsically complex and need to be described with many features, and when at the same time they are available only in potentially very limited quantities. Formally, a dataset $T$ composed of $\{(\mathbf{x}_{1}, y_{1}), (\mathbf{x}_{2}, y_{2}), \ldots, (\mathbf{x}_{n}, y_{n})\}$ instances, where $\mathbf{x}_i$ is the vector of descriptive features belonging to $\mathbb{R}^{m}$ and $y_i$ is its corresponding class label, is considered to be a HDLSS dataset when $m \gg n$ \cite{review:2014, qingbo:2020}. However, it should be noted that there is no consensus on the threshold to apply to the ratio between $m$ and $n$ to unambiguously decide whether a dataset is pertaining to an HDLSS learning problem \cite{review:2014, qingbo:2020, Nakayama:2019}.

HDLSS datasets are recurring in many real-world applications, including, but not limited to, medical imaging, DNA microarrays, and text processing \cite{review:2014, svm_feat_sel}. For instance, for pattern recognition tasks in medical imaging, numerical image representations are known to be high dimensional, whether they are built with hand-crafted features or with automatically learned deep features \cite{Ma:2019}. This is combined with the fact that acquiring samples for medical-related pattern recognition applications is not an easy task due to health data privacy policies, political concerns, or even the homogenization of medical protocols. In general, this means dealing with particularly small-sized datasets.

HDLSS datasets pose a number of challenges to general-purpose machine learning algorithms. These datasets usually embed very complex concepts due to a large number of relevant features, while they generally do not contain a sufficient number of instances for learning these concepts. This leads to several machine learning issues, well known to the community and often referred to as the curse of dimensionality. For example, for the many methods based on distance metrics, like the $k$-Nearest Neighbors methods, the notion of ``neighborhood'' becomes progressively ill-defined in high dimensions \cite{Gunduz:2015}. This is because the distances between instances become more and more similar when the dimension increases \cite{Kouiroukidis:2011}. Another example of difficulties encountered in learning HDLSS data is the presence of outliers. Some methods are known to be particularly sensitive to outliers \cite{Zhang:2011}, and such data are quite common in high-dimensional feature spaces \cite{Gunduz:2015}. A last example of well-known learning problems often encountered in the HDLSS context is overfitting. In this context, many general-purpose machine learning techniques are likely to overfit. It is the case of SVM-based methods for example, for which the so-called data-piling phenomenon is particularly salient in the HDLSS context and leads to strong overfitting \cite{marron:2007}.

The most common approach in the literature for dealing with HDLSS learning tasks is based on dimensionality reduction. The key idea is to reduce the dimension to transform a high dimensional problem into a learning problem where $m \approx n$. However, these methods can be considered as workarounds rather than true solutions for learning HDLSS datasets. Moreover, they often suffer from two major drawbacks:
\begin{itemize}
    \item Using dimensionality reduction often result in a significant loss of information if the features are mostly relevant and poorly correlated \cite{cao:2019, Shen:2022}.
    \item Selecting a small subset of the most relevant features may not yield good results if the number of data is too small \cite{Kuncheva:2020, Shen:2022}.
\end{itemize}

In contrast, few methods have been proposed in the literature specifically to handle HDLSS learning tasks, which we review in Section \ref{sec:rel_work}.
The most efficient ones are derived from the SVM principle, with a modified formulation of the underlying optimization problem in order to better adapt to the HDLSS specificities \cite{Shen:2020, Shen:2022}.
However, we think that another promising approach is to use a HDLSS-compliant similarity measure as a kernel for SVM, instead of relying on a specific problem formulation. This idea has been recently applied to multi-view learning as in this context, similarity representations allow to easily merge the different views together \cite{cao:2019}. This approach, named Random Forest Dissimilarity (RFD), leans on Random Forest classifiers to build dissimilarity representations that are then used as pre-computed kernels in SVM classifiers. We show in this paper that this principle can be straightforwardly and efficiently applied to HDLSS classification problems, for which we think it presents real assets: (i) Learning based on (dis)similarities between instances is a good way to deal with particularly small-sized datasets and (ii) the Random Forest (dis)similarity measure is known to be particularly robust to high dimensions. Therefore, this work presents:
\begin{itemize}
    \item a transposition of the RFD method to HDLSS classification tasks with an emphasis on its strength to face the HDLSS challenges.
    \item a rigorous experimental validation, including comparisons with several state-of-the-art HDLSS learning methods on 40 real-world problems, along with a thorough statistical analysis of the results.
\end{itemize}
Note that for simplicity, we focus only on classification problems in this study. However, our proposal is straightforwardly applicable to regression tasks, as well as all the other methods used for comparison.

The remainder of this paper is organized as follows. Section \ref{sec:rel_work} reviews the main state-of-the-art approaches for HDLSS learning. Section \ref{sec:rfdis} presents the Random Forest SVM method and discuss its assets fro HDLSS classification. Section \ref{sec:xp} describes the experimental setting, followed by the presentation and analysis of the results in Section \ref{sec:res}. Finally, Section \ref{sec:conclusion} gives our conclusion and future works.

\section{Related Work}
\label{sec:rel_work}

This section focuses on solutions to address the challenges posed by HDLSS learning tasks and provides an overview of the leading solutions that can be found in the literature, with the exception of dimensionality reduction approaches, not included for reasons explained in the introduction.

\subsection{Limitations of traditional methods for dealing with HDLSS problems}

Traditional general-purpose machine learning techniques like Discriminant Analysis \cite{Freidman1989}, $k$-Nearest Neighbors \cite{dutta:2016}, and Support Vector Machines \cite{marron:2007} usually fail to handle HDLSS datasets mainly due to the fact that such a context leads to ill-posed problems or to unsuitable learning conditions. For example, Linear Discriminant Analysis (LDA) suffers from a well-known problem when the dimension is larger than the number of training instances, i.e., $m \gg n$. The underlying principle of LDA is to find a projection of the data in which the between-class separability is maximized while the within-class variability is minimized. To do so, it uses a within-class scatter matrix that is known to be singular when $m \gg n$. This is an important problem since the non-singularity of this matrix is required to find the LDA basis vectors. The usual ways to circumvent this situation is either to perform dimensionality reduction beforehand (which is beyond the scope of this work as previously explained), or to use regularization techniques, as in the Regularized Discriminant Analysis \cite{Freidman1989} which has been successfully used for real-world HDLSS problems \cite{Guo2006}.

Distance-based classification techniques, like the $k$-Nearest Neighbors family of methods, are classifiers that usually perform well when $n > m$. However, they are also known to suffer from the curse of dimensionality in the opposite situation because the pairwise distances between all observations concentrate around a single value in this case \cite{Francois2007}. Similarly as before, a large part of the solutions proposed in the literature are based on dimensionality reduction techniques, e.g., \cite{Deegalla2006}. A few others, however, propose alternative mechanisms for dealing with HDLSS data like weighted voting scheme \cite{Radovanovic2009}, fuzzy neighborhood \cite{Tomasev2011}, or new proximity measurements \cite{Pal2016} to name a few.

Nevertheless, most state-of-the-art methods tailored for HDLSS classification in the literature are based on adaptations of SVM classifiers. SVM are known to be particularly prone to overfitting in the HDLSS context, which is often illustrated through the data-piling phenomenon \cite{marron:2007}. 
The following section details this phenomenon as well as the different solutions that have been proposed in the literature.

\subsection{SVM-inspired methods}

In binary classification, the underlying principle of Support Vector Machines (SVM) \cite{Cortes:1995} is to find a hyperplane that best separates the instances according to their classes by maximizing the distance (called the margin) to its closest instances (called the support vectors). 
The solution is the hyperplane whose parameters that minimizes the following problem:
\begin{align}
    \label{eq:svm}
        \min_{\mathbf{w},b,\{\xi_i\}} &\frac{1}{2} \Vert \mathbf{w} \Vert^2 + C \sum_{i=0}^{n} \xi_i \\
        \textrm{s.t.} \ &y_{i}\left(\mathbf{w}^\top \mathbf{x}_i + b\right) \geq 1-\xi_i, i=1,\dots,n  \nonumber \\
        & \xi_i \geq 0, i=1,\dots,n  \nonumber
\end{align}
where $\mathbf{w}$ is the normal vector of the hyperplane, $b$ its intercept term, and where $C$ is a regularization hyperparameter to control a trade-off between maximizing the margin and taking into account some training errors when the problem is not strictly linearly separable.

The behavior of this SVM method for HDLSS classification has been extensively analyzed and discussed in \cite{marron:2007}. The authors show that in such a setting, SVM classifiers face the so-called \textit{data-piling} problem. This problem arises specifically with HDLSS datasets because, in this case, a large proportion of the training instances are support vectors, i.e. they lie on the margin boundaries  
\rev{resulting from the minimization of Equation \ref{eq:svm}. When projected in the discriminant direction (i.e. to the normal vector $\mathbf{w}$ obtained by minimizing Equation \ref{eq:svm}), all these support vectors "pile up on top of each other", i.e. they are projected onto exactly two points, one for each class. In this discriminant projection, the separating hyperplane is exactly halfway between these two points. Nevertheless, this usually reflects severe overfitting since independent test instances may not be projected the same way in the discriminant direction. As a result, while the resulting linear classifier perfectly fit (most of) the training instances, there is a strong risk that it will not generalize well to new data points. This phenomenon has been widely illustrated and analyzed in the literature and we refer the reader to \cite{marron:2007, qiao:2010, qingbo:2020} for further details.}

Variations have been proposed to solve this problem, mainly by modifying the underlying optimization problem. For example, the Distance Weighted Discrimination (DWD) \cite{marron:2007} leans on a minimization problem for which the best separating hyperplane is the one that maximizes the harmonic mean of all distances to the hyperplane:
\begin{align}
    \min_{\mathbf{w},b,\{\xi_i\}} &\sum_{i=1}^{n} \left(\frac{1}{r_{i}} + C \xi_i \right)  \\
    \textrm{s.t.} \ &r_{i} = y_i\left( \mathbf{w}^\top\mathbf{x}_i + b\right) + \xi_i, \nonumber \\
        & r_{i} \geq 0, \xi_i \geq 0, \Vert \mathbf{w} \Vert^2 \leq 1 \nonumber
\end{align}

In this process all training instances are taken into account to find the hyperplane, instead of relying only on the support vectors. A variant of this principle, named Weighted Distance Weighted Discrimination (wDWD) \cite{qiao:2010}, has been proposed to allow for more flexibility and robustness, as the method is known to be sensitive to imbalanced classes and quickly become computationally expensive as the number of training instances increases \cite{qingbo:2020}.

Another SVM-inspired method is proposed in \cite{qingbo:2020} named the Population-Guided Large Margin Classification (PGLMC) method, that aims to take up the idea of DWD while improving its computational performance. Equation \ref{eq:svm} is modified to take into account the distance between the centroids of the classes (instead of all the traning instances) to ensure the training instances of both classes are as far as possible along the projecting direction $\mathbf{w}$:
\begin{align}
    \min_{\mathbf{w},b,\{\xi_i\}} &\frac{\Vert \mathbf{w} \Vert^2}{(m_1 - m_2)^{\top}\mathbf{w}} + C \sum_{i=0}^{n} \xi_i \\
    \textrm{s.t.} \ &y_{i}\left(\mathbf{w}^\top \mathbf{x}_i + b\right) \geq 1-\xi_i, i=1,\dots,n  \nonumber \\
    & (m_1 - m_2)^{\top}\mathbf{w} \geq 2 \\
    & \xi_i \geq 0, i=1,\dots,n  \nonumber
\end{align}
where $m_1$ (resp. $m_2$) is the centroid of the first class (resp. the second class). Following a similar principle, a slightly different formulation is proposed in \cite{Shen:2022} that still maximize the distance between classes but that also strives to distribute the points as much as possible in the projection direction to avoid data-piling. The resulting method is named No-separated Data Maximum Dispersion classifier (NPDMD).

\subsection{SVM with an HDLSS-robust kernel}


\rev{The solutions mentioned above are all based on different formulations of the underlying optimization problem, whose solution leads to the final linear classifier. However, SVMs are known to allow the learning of non-linear classifiers using the kernel trick. Intuitively, it consists in applying a non-linear projection of the data into an \textit{implicit} feature space in which a separating hyperplane exists. The key feature of this projection is that it does not require coordinates to be calculated explicitly, as all operations are performed through scalar products in this space, calculated with a kernel function. 

Formally, this trick is based on the Lagrangian dual form of the SVM optimization problem:
\begin{align}
    \label{eq:svmdual}
        \max_{\bm{\alpha}} \quad & \sum_{i=1}^{n} \alpha_i - \frac{1}{2} \sum_{i=1}^{n} \sum_{j=1}^{n} \alpha_i \alpha_j y_i y_j \mathbf{x}_j^\top\mathbf{x}_i \\
        \textrm{s.c.} \quad & \alpha_i \geq 0, \quad i=1,\dots,n \\
        & \sum_{i=1}^{n} \alpha_i y_i =0
\end{align}
where the $\alpha_i$ are the Lagrange multipliers. This dual form can be efficiently solved by quadratic programming algorithm, with the notable advantage of searching for $n$ parameters (the Lagrange multipliers) instead of $m+1$ parameters with the primal version (the $m$ values of $\mathbf{w}$ and $b$). This makes it particularly suitable for HDLSS problems where $m \gg n$. This also allows to express the resulting classifier has a function of the support vectors:
\begin{equation}
    \label{eq:dual_classif}
    h(\mathbf{x}) = \sum_{i=1}^{n} \alpha_i y_i \mathbf{x}_i^\top \mathbf{x} + b = \sum_{i=1}^{n} \alpha_i y_i K(\mathbf{x}_i,\mathbf{x}_j) + b
\end{equation}
As a consequence, all instances are only accessed through scalar product, allowing the use of any kernel functions $K$, as in the right-hand expression of Equation \ref{eq:dual_classif}. We refer the reader to \cite{Hofmann:2008} for further details about the kernel trick, and kernel methods in general.
}

There are several popular kernels for SVM in the literature, but the best performing one for a wide range of real-world problems is the radial basis function (RBF) kernel, defined as:
\begin{equation}
    K(\mathbf{x}_i,\mathbf{x}_j) = \exp \left(-\gamma \Vert \mathbf{x}_i - \mathbf{x}_j \Vert^2 \right)
\end{equation}
\noindent where $\mathbf{x}_i,\mathbf{x}_j$ is any pair of instances, and where $\gamma$ is a hyperparameter. It is important to note that the behavior of the resulting model is highly sensitive to the value of $\gamma$ and that it must be tuned in conjunction with the regularization parameter $C$. The traditionally recommended protocol for tuning these hyperparameters is detailed in Section \ref{sec:xp}.

The asymptotic behaviors of SVM with RBF kernel in the HDLSS context are further investigated in \cite{Nakayama:2019}. This study shows that nonlinear SVM classifiers with RBF kernels are highly biased in the HDLSS context, especially with imbalanced classes, and we therefore believe that using an HDLSS-robust kernel would be more relevant. For this, one can rely on the fact that the kernels can be directly interpreted as a similarity measurement \cite{vert:2004}. In fact, any similarity measure can be used as kernel provided that it fulfills specific mathematical conditions \cite{pekalska:2002,cao:2019}. Therefore, using SVM for tackling the HDLSS classification challenges could be addressed by proposing a suitable similarity measure instead of modifying the underlying SVM optimization problem. This is the main idea of the method we propose to evaluate in this work and which is described in the next section.

\section{The RFSVM method \label{sec:rfdis}}

\subsection{Using Random Forest as a kernel}

Random Forest is a very versatile general-purpose learning method, that have shown to be accurate for many real-world problems \cite{Delgado:2014}. These methods are also known to provide a number of mechanisms for analysis and interpretability, such as 
a similarity (or proximity) measure \cite{cao:2019}. 

For computing the RF similarity between any pair of instances, one must have a previously trained RF classifier noted $H(\mathbf{x}) = \{ h_k(\mathbf{x}) ~\vert ~ 1 \leq k \leq M\}$, made up with $M$ decision trees $h_k$. Any RF learning algorithm can be used for that purpose, as the similarity measurement leans on the final ensemble of decision trees grown during learning.
\rev{Note that for all experiments of this work, the Breiman's Random Forest method \cite{Breiman:2001} has been used, via the implementation proposed in the Scikit-learn python library \cite{pedregosa:2011}.} 
The similarity between two instances ($\mathbf{x}_i,\mathbf{x}_j$) is inferred by the forest by comparing the descending paths followed by both instances in each tree: let $\mathcal{L}_k$ denote the set of leaves of $h_k$, and let $l_k(\mathbf{x})$ denote a function from the input domain $\mathcal{X}$ to $\mathcal{L}_k$, that returns the leaf of $h_k$ where $\mathbf{x}$ lands when one wants to predict its class. The similarity measure $s_k$ is defined as in Equation \ref{eq:simDT}: if the two instances $\mathbf{x}_i$ and $\mathbf{x}_j$ land in the same leaf of $h_k$, then the similarity between both instances is set to $1$, else it is equal to $0$.
\begin{equation}
    \label{eq:simDT}
    s_k(\mathbf{x}_i, \mathbf{x}_j) = 
    \left\{ 
        \begin{array}{ll}
            1, & \text{if} \  l_k(\mathbf{x}_i)=l_k(\mathbf{x}_j) \\
            0, & \text{otherwise}
        \end{array}
    \right.
\end{equation}

The RF similarity measure $s_H(\mathbf{x}_i, \mathbf{x}_j)$ derived from the whole forest $H$ consists in calculating $s_k$ for each tree $h_k$, and in averaging the resulting values over the $M$ trees:
\begin{equation}
    \label{eq:dissRF}
    s_H(\mathbf{x}_i, \mathbf{x}_j) = \frac{1}{M} \sum_{k=1}^{M} s_k(\mathbf{x}_i, \mathbf{x}_j)
\end{equation}

A remarkable property of this similarity measure is that it can be assimilated to a positive semi-definite (p.s.d.) kernel function \cite{cao:2019}. Similarly, one can show that the similarity matrix $\mathbf{S}_H$:
\begin{equation}
    \mathbf{S}_H =
    \begin{bmatrix}
        s_H(x_1, x_1) & s_H(x_1, x_2) & \ldots & s_H(x_1, x_n)\\
        s_H(x_2, x_1) & s_H(x_2, x_2) & \ldots & s_H(x_2, x_n)\\
        \ldots & \ldots & \ldots & \ldots\\
        s_H(x_n, x_1) & s_H(x_n, x_2) & \ldots & s_H(x_n, x_n)\\
    \end{bmatrix}
\end{equation}
whose elements are the RF similarity measures between each pair of training instances, is a positive semi-definite matrix (see the proof in Appendix A of \cite{cao:2019}). As a consequence, such a matrix can be used in a kernel method as a pre-computed kernel. 

In a nutshell, using such a pre-computed kernel with SVM classifier, noted RFSVM in the following, consists of:
\begin{enumerate}
    \item Training a Random Forest classifier $H$ on the training set $T$;
    \item Building a similarity matrix $\mathbf{S}_H$ from $H$;
    \item Feeding an SVM classifier with $\mathbf{S}_H$ as a pre-computed kernel.
\end{enumerate}
As for the prediction phase, once the RFSVM classifier is trained this way, an unseen testing instance $\mathbf{x}$ can be predicted as follows:
\begin{enumerate}
    \item The similarity values between $\mathbf{x}$ and all the $n$ training instances from $T$ are computed, leading to a $n$-sized vector of $s_H(\mathbf{x}, \mathbf{x}_i), i=1,\dots,n$;
    \item This vector is given as input to the RFSVM classifier for prediction.
\end{enumerate}
\rev{The whole learning procedure is further detailed in Algorithm \ref{algo:rfsvm}. Note that the Random Tree building method used in this work is the CART-based implementation from the Scikit-learn library \cite{pedregosa:2011} and the SVM solver used in this work is the LIBSVM solver that implements the sequential minimal optimization (SMO) algorithm for kernelized SVM \cite{Platt:1998}.}

\rev{
\begin{algorithm2e}[hbt!]
    \footnotesize
	\DontPrintSemicolon
	\SetKwInOut{Input}{input}
	\SetKwInOut{Output}{output}
	\caption{The RFSVM learning method\label{algo:rfsvm}}
	
	\Input{$T$, a training set composed of $n$ instances $(\mathbf{x}_{i}, y_{i})$}
	\Input{$M$, the number of trees in the random forest}
	\Input{$\Theta$, the random tree learning hyperparameter values}
	\Input{$C$, the SVM regularization hyperparameter value}
	\Output{$SVM_H$, a RFSVM model trained on $T$}
	\BlankLine

	\Begin{
		\BlankLine
		\tcp{1. Train a random forest composed of $M$ random trees}
		$H \gets \emptyset$ \;
		\For{$k=0$ \KwTo $M$}{
            $T_k \gets BootstrapSampling(T)$ \;
			$h_k \gets RandomTree(T_k, \Theta)$ \;
			$H \gets H \cup h_k$ \;
		}
		\tcp{Note: the booststrap sampling method and the random tree method used in this work are those of the random forest implementation of the Scikit-learn library \cite{pedregosa:2011}}
		\BlankLine
		\tcp{2. Compute the similarity matrix from the random forest}
		$S_H \gets I_n$ \tcc*[r]{init. $S_H$ to a $n \times n$ identity matrix}
		\For(\tcc*[f]{for each pair $\mathbf{x}_i,\mathbf{x}_j \in T$}){$i=1$ \KwTo $n-1$}{
			\For{$j=i+1$ \KwTo $n$}{
				\For(\tcc*[f]{for each tree $h_k \in H$}){$k=1$ \KwTo $M$}{
					$l_i \gets getLeaf(h_k,\mathbf{x}_i)$ \tcc*[r]{get the leaf of $h_k$ where $\mathbf{x}_i$ lands}
					$l_j \gets getLeaf(h_k,\mathbf{x}_j)$ \tcc*[r]{get the leaf of $h_k$ where $\mathbf{x}_j$ lands}
					\If{$l_i = l_j$}{	
						$S_H(i,j) \gets S_H(i,j) + 1$ \;
					}
				}
				$S_H(i,j) \gets S_H(i,j) / M$ \;
				$S_H(j,i) \gets S_H(i,j)$ \;
			}
		}
		\BlankLine
		\tcp{3. Use $\mathbf{S}_H$ as a precomputed kernel for the SVM learning}
		$SVM_H \gets LIBSVMSolver(\mathbf{S}_H, C)$ \;
		\tcp{Note: the LIBSVM solver used in this work implements the sequential minimal optimization (SMO) algorithm \cite{Platt:1998}}
	}
\end{algorithm2e}
}

\subsection{Discussion}

It is worth noting that the RFSVM procedure can be applied with any similarity measure provided that the resulting similarity matrix is p.s.d. 
For example, the well-known cosine measure could be used to replace the Random Forest similarity measure. 
However, we would like to stress that the Random Forest similarity measure is particularly suitable to do so for classification tasks. 
The reason is that it is computed in such a way it can well reflect the class belonging of the instances. 
According to this measure, two instances that belong to the same class are more likely to be similar than two instances from different classes.
This is due to the fact that the Random Forest classifier is built beforehand by taking the class into account so that the leaves of all the trees are expected to gather instances from the same class.

Similarly, metric learning methods could replace the RF classifier for inferring the similarity values. 
However, we argue that the state-of-the-art metric learning methods require a formulation of the metric beforehand, which is not the case of RF methods, and more importantly, they are known to be sensitive to high dimensions. 
For example, most of the state-of-the-art metric learning methods that are based on the Mahalanobis distance suffer from two important drawbacks for HDLSS classification:
\begin{enumerate}
    \item They are computationally intractable in high dimensions since the number of parameters to learn (the covariance matrix elements) is $\mathcal{O}(m^2)$.
    \item They face a strong risk of overfitting phenomenon since the number of training instances used to estimate these parameters is very low in the HDLSS context.
\end{enumerate}

Consequently, we argue that classical metric learning methods are not suitable for HDLSS problems, contrary to RF classifiers, known to be very robust to high dimensions without the need for a proportionally large training set. 
To support this statement, the most popular metric learning method, namely Large Margin Nearest Neighbors (LMNN) \cite{Domen2005}, has been included in the experiments as an alternative to RF classifiers for building a pre-computed kernel in SVM. 
The resulting method is named LMNNSVM in the following.

The RFSVM method explained in the previous section has been successfully applied to multi-view classification problems \cite{cao:2018,cao:2019,cao:2020}. 
These studies have mainly focused on using RF (dis)similarity matrices on each view and fuse them in order to take benefit from the complementarity between the different views.
However, the extent to which the success of this approach depends on how well it exploits complementarities in multi-view learning or on the use of the RF similarity measure itself has not yet been studied. Therefore, the main contribution of this work is to extend the experimental study and validate this approach to regular HDLSS single-view classification problems.

\section{Experiments \label{sec:xp}}

This section details the experimental protocol, as well as its underlying goal of evaluating how the proposed method compares to general-purpose machine learning methods, similarity-based methods, and HDLSS-specific methods for a variety of datasets that exhibit different levels of HDLSS.

\subsection{Datasets}

Our experimentation encompasses 40 public datasets, among which 21 datasets were acquired from the OpenML repository \cite{vanschoren:2013}, 3 text processing datasets from the UCI repository \cite{dua:2019}, and 16 medical datasets from \cite{souto:2008}. 

In the literature, the only consensus to formally define an HDLSS problem is ``$m \gg n$'' \cite{Nakayama:2019,qingbo:2020,review:2014}. 
\rev{Therefore, \textit{HDLSSness} could be measured by the ratio between the number of instances and the number of features in the associated dataset. However, this does not take into account the number and imbalance of classes, which can have a major impact on the difficulties arising from HDLSS learning. }
Therefore, we propose to quantify a level of HDLSS for each dataset, with the measure defined as:
\begin{equation}
    \label{eq:hdlssness}
    \begin{aligned}
        \Omega = \frac{1}{m}\times\frac{\sum_{j=1}^{c}{n_j}}{c},
    \end{aligned}
\end{equation}
where $c$ is the number of classes, $m$ is the number of features, and $n_j$ is the number of instances from the $j$-th class in a dataset. This $\Omega$ measure corresponds to the average number of instances per class divided by the number of features in a dataset. Consequently, the smaller the value of $\Omega$, the more HDLSS a dataset is. 
\rev{Using the average number of instances per class, instead of the total number of instances in the dataset, allows to limit the bias introduced by strong class imbalances, such as those found in some of the datasets we selected in our experiments.}

Table \ref{tab:datasets} gives a description of all the datasets used in our experiments, with the number of instances, the number of features, the imbalance ratio (IR), the number of classes, and the $\Omega$ value. The IR is computed by dividing the number of instances from the majority class by the number of instances from the minority class. In this table, datasets are sorted by increasing value of $\Omega$. One can observe that as $\Omega$ increases, the number of instances also increases, and the dimensionality decreases. This table also contains a separation between HDLSS datasets in the upper part and non-HDLSS datasets in the lower part, corresponding to values of $\Omega = 1$. This allows to highlight that these datasets have been chosen in a wide range of cases, including datasets corresponding to traditional classification problems. This will also allow us to show that the RFD method remains competitive in a more classical learning context.

\begin{table}
    \centering
    \caption{ \label{tab:datasets}
    Datasets description with the number of instances, the number of features, the imbalance ratio (IR), the number of classes, and the $\Omega$ value. Datasets with $\Omega < 1$ are considered to be HDLSS datasets, whereas $\Omega \geq 1$ are non-HDLSS datasets.}
    \begin{tabular}{|c c c c c c|}
        \hline
        Name & Instances & Features & IR & Classes & $\Omega$  \\ [0.5ex]
        \hline\hline
        UMIST\_Faces\_Cropped \cite{vanschoren:2013} & 575 & 10304 & 2.526 & 20 & 0.003\\
        \hline
        leukemia \cite{vanschoren:2013} & 72 & 7129 & 1.88 & 2 & 0.005\\
        \hline
        alizadeh-2000-v3 \cite{souto:2008} & 62 & 2091 & 2.333 & 4 & 0.007\\
        \hline
        tr45.wc \cite{vanschoren:2013} & 690 & 8261 & 11.429 & 10 & 0.008\\
        \hline
        laiho-2007 \cite{souto:2008} & 37 & 2202 & 3.625 & 2 & 0.008\\
        \hline
        bittner-2000 \cite{souto:2008} & 38 & 2201 & 1.0 & 2 & 0.009\\
        \hline
        arcene \cite{vanschoren:2013}& 200 & 10000 & 1.273 & 2 & 0.010\\
        \hline
        ramaswamy-2001 \cite{souto:2008} & 190 & 1363 & 3.0 & 14 & 0.010\\
        \hline
        armstrong-2002-v2 \cite{souto:2008} & 72 & 2194 & 1.4 & 3 & 0.011\\
        \hline
        su-2001 \cite{souto:2008}& 174 & 1571 & 4.667 & 10 & 0.011\\
        \hline
        lapointe-2004-v2 \cite{souto:2008}& 110 & 2496 & 3.727 & 4 & 0.011\\
        \hline
        golub-1999-v2 \cite{souto:2008}& 72 & 1868 & 4.222 & 3 & 0.013\\
        \hline
        Dexter \cite{vanschoren:2013} & 600 & 20000 & 1.0 & 2 & 0.015\\
        \hline
        yeoh-2002-v2 \cite{souto:2008}& 248 & 2526 & 5.267 & 6 & 0.016\\
        \hline
        tomlins-2006-v2 \cite{souto:2008}& 92 & 1288 & 2.462 & 4 & 0.018\\
        \hline
        khan-2001 \cite{souto:2008}& 83 & 1069 & 2.636 & 4 & 0.019\\
        \hline
        west-2001 \cite{souto:2008}& 49 & 1198 & 1.042 & 2 & 0.020\\
        \hline
        eating \cite{vanschoren:2013} & 945 & 6373 & 1.176 & 7 & 0.021\\
        \hline
        bhattacharjee-2001 \cite{souto:2008}& 203 & 1543 & 23.167 & 5 & 0.026\\
        \hline
        micro-mass \cite{vanschoren:2013} & 360 & 1300 & 1.0 & 10 & 0.028\\
        \hline
        oh15.wc \cite{vanschoren:2013} & 913 & 3100 & 2.962 & 10 & 0.029\\
        \hline
        oh10.wc \cite{vanschoren:2013} & 1050 & 3238 & 3.173 & 10 & 0.032\\
        \hline
        shipp-2002-v1 \cite{souto:2008}& 77 & 798 & 3.053 & 2 & 0.048\\
        \hline
        cnae-9-half \cite{vanschoren:2013} & 540 & 856 & 1.283 & 9 & 0.070\\
        \hline
        OVA\_Colon \cite{vanschoren:2013} & 1545 & 10935 & 4.402 & 2 & 0.071\\
        \hline
        OVA\_Breast \cite{vanschoren:2013} & 1545 & 10935 & 3.491 & 2 & 0.071\\
        \hline
        imdb \cite{dua:2019} & 748 & 3047 & 1.066 & 2 & 0.123\\
        \hline
        cnae-9 \cite{vanschoren:2013} & 1080 & 856 & 1.0 & 9 & 0.140\\
        \hline
        lsvt \cite{vanschoren:2013} & 126 & 310 & 2.0 & 2 & 0.203\\
        \hline
        yelp \cite{dua:2019}& 1000 & 2033 & 1.0 & 2 & 0.246\\
        \hline
        amazon \cite{dua:2019} & 1000 & 1847 & 1.0 & 2 & 0.271\\
        \hline
        chowdary-2006 \cite{souto:2008}& 104 & 182 & 1.476 & 2 & 0.286\\
        \hline
        \hline
        chen-2002 \cite{souto:2008}& 179 & 85 & 1.387 & 2 & 1.053\\
        \hline
        gina \cite{vanschoren:2013} & 3153 & 970 & 1.034 & 2 & 1.625\\
        \hline
        madelon \cite{vanschoren:2013} & 2600 & 500 & 1.0 & 2 & 2.600\\
        \hline
        scene \cite{vanschoren:2013} & 2407 & 299 & 4.585 & 2 & 4.025\\
        \hline
        wdbc \cite{vanschoren:2013}& 569 & 30 & 1.684 & 2 & 9.483\\
        \hline
        led24 \cite{vanschoren:2013}& 3200 & 24 & 1.139 & 10 & 13.333\\
        \hline
        segment \cite{vanschoren:2013}& 2310 & 19 & 1.0 & 7 & 17.368\\
        \hline
        spambase \cite{vanschoren:2013}& 4601 & 57 & 1.538 & 2 & 40.360\\
        \hline
    \end{tabular}
\end{table}

\subsection{Methods and parametrization}

In addition to the RFSVM method, several learning methods were selected for comparison, in three families of methods: general-purpose methods, SVM variants, and similarity-based methods. 
In the first group, the Random Forest classifier and the Extreme Gradient Boosting (XGBoost) method were selected because of their state-of-the-art performance on various ML problems \cite{Breiman:2001, Chen:2016, Delgado:2014}.
Regarding SVM variants, the regular method with RBF kernel \cite{Cortes:1995} has been retained, as well as the DWD method presented in Section \ref{sec:rel_work}. The reasons why the DWD methods has been retained in our experiments instead of the other SVM variants listed in Section \ref{sec:rel_work} are (i) because the results obtained in \cite{Shen:2022} from the experimental comparison between all these methods show very similar results without any statistical test of significance supporting the superiority of one method over the others and (ii) because it is the only one with a freely available implementations. As for similarity-based methods, the cosine distance and the LMNN variant of the Mahalanobis distance were used in the same way as with the RFSVM method, i.e. as a precomputed kernel, to support the claims given in the discussion subsection of the Section \ref{sec:rfdis}.

For each dataset, all classifiers had their hyperparameters tuned using a 3-fold cross-validation procedure. The tuning process was performed using the \texttt{hyperopt} library \cite{bergstra:2013}. This library requires as input a search space for each hyperparameter and the number of evaluations for the tuning process. In the following experiments, the number of evaluations was set to 100, and the optimization criterion was accuracy maximization. Internally, \texttt{hyperopt} conducts hyperparameter optimization by converting the search process into a generative process using Tree-structured Parzen Trees (TPEs) \cite{NIPS2011_TPE}. We refer the reader to \cite{bergstra:2013} for more details about its functionning.

Regarding the Random Forest method, the maximum tree depth\footnote{The \texttt{None} option stands for fully grown trees, i.e., the trees are grown until all leaf nodes have pure class distributions.}, the maximum number of features assessed for a split decision, the minimum number of samples at leaf nodes, the minimum number of samples for a split, and the number of trees have been tuned following the values given in the upper part of Table \ref{tab:hp}. The search space for XGBoost is given in the middle part of the same table. XGBoost search space follows the suggestion given in \cite{putatunda:2018}, where the maximum tree depth, instance subsample ratio, column sample by tree portion, regularization lambda, and the maximum number of iterations were tuned. Finally, the search space for SVM with the gaussian kernel is depicted in the bottom part of Table \ref{tab:hp}, where the hyperparameter \(C\) and \(\gamma\) were optimized. For all other SVM-based methods, i.e. RFSVM, DWD, COSSVM and LMNNSVM, only $C$ needs to be optimized, and it has also been done following the values in Table \ref{tab:hp}. 

\begin{table}[!h]
    \centering
    \caption{  \label{tab:hp} Hyperparameter search space for Random Forest, XGBoost, SVM, and other SVM-based methods (RFSVM, COSSVM, LMNNSVM, and DWD).}
    \begin{tabular}{|l c|}
        \hline
        \multicolumn{2}{|c|}{\textbf{Random Forest}} \\
        \hline
        Max. Depth & $\{10^i \vert i=1,\ldots,10\}$ and \texttt{None} \\
        \hline
        Max. Features & \{1\%, 5\%, 10\%, 20\%, 30\%\} \\
        \hline
        Min. Samples Leaf & \{1, 2, 4\} \\
        \hline
        Min. Samples Split & \{2, 5, 10\} \\
        \hline
        Number of trees & 500 \\
        \hline\hline
        \multicolumn{2}{|c|}{\textbf{XGBoost}} \\
        \hline
        Max. Depth &	$\{x \vert x \in \mathbb{N}, 4 \leq x \leq 15\}$\\
        \hline
        Subsample & $[0.8,1]$  \\
        \hline
        Column sample by tree & $[0.5,1]$ \\
        \hline
        Regularization Lambda & $[0,1]$\\
        \hline
        Max. number of iterations ~~~& 500\\
        \hline\hline
        \multicolumn{2}{|c|}{\textbf{SVM}} \\
        \hline
        C & $\{10^i \vert~ i=-2,\dots,4 \}$\\
        \hline
        $\gamma$ & $\{10^i \vert~ i=-4,\dots,2 \}$ \\
        \hline\hline
        \multicolumn{2}{|c|}{\textbf{RFSVM, COSSVM, LMNNSVM, and DWD}} \\
        \hline
        C & $\{10^i \vert~ i=-2,\dots,4 \}$\\
        \hline
    \end{tabular}
\end{table}

\subsection{Validation protocol and implementation details}

For each dataset used in this experimental comparison, all the methods were tested 10 times, each time with a random half of the dataset for training and the remaining half for test. The hyperparameter setting procedure was performed on the training set and the performance was measured with the traditional accuracy measurement.

Two different statistical tests of significance have been used to analyze the differences in performance between the methods. The tests used are (i) the Friedman test along with the Nemenyi post-hoc test \cite{demsar:2006}, and (ii) the Bayesian sign test \cite{Alessio:2017}. The Friedman/Nemenyi test is classically used to assess the statistical significance of a comparison of several methods over multiple datasets, based on the average ranks of the methods. We refer the reader to \cite{demsar:2006} for more details about its functioning and the reasons why this test is advised for this type of comparisons. In contrast, the Bayesian sign test is a pairwise test based on the difference in performance of the two methods over multiple datasets. Unlike frequentist null-hypothesis tests such as the Friedman test, Bayesian analysis can provide more insight than simply rejecting the hypothesis that the two classifiers are of equivalent performance. In particular, it outputs probabilities of one classifier $a$ to be practically better than another classifier $b$, based on a set of results from multiple datasets. It also allows to integrate in the analysis a ``region of practical equivalence'' (\textit{rope}) defined by an average performance gap at which the classifier $a$ is considered practically better than the classifier $b$. For instance, with \textit{rope} = 0.05, if the mean accuracy of classifier \(a\) is $0.980$ and the mean accuracy of classifier \(b\) is $0.976$, both classifier will be considered to be practically equivalent in the Bayesian analysis because the difference $0.980 - 0.976 = 0.04$ is less than the \textit{rope}. In our experiments, we considered two values for \textit{rope}, $0.005$ and $0.01$.

Finally, all the experiments were executed in Python. Random Forest (RF), SVM, and cosine distance implementations were those available in the \textit{scikit-learn} library \cite{pedregosa:2011} version 0.23.2. The implementation of DWD was found in the IDC9 repository\footnote{https://github.com/idc9/dwd}, and the PYLMNN library \cite{weinberger:2009} version 1.6.4 was used for computing the LMNN distance. XGBoost was performed using the implementation provided in its original paper (version 1.1.1).

\section{Results and Discussion \label{sec:res}}

Table \ref{tab:results_all_hdlss} presents the mean accuracy rates (and standard deviations) obtained by the seven methods across all 40 datasets. The values in bold are the best mean accuracy rates obtained for each of the datasets. In this table, the datasets are separated into three groups: the very-HDLSS datasets at the top (when $\Omega < 0.015$), the mid-HDLSS datasets in the middle (when $ 0.015 \leq \Omega < 1$), and non-HDLSS datasets at the bottom. In this section, we analyze these results first globally and then in more detail for each of these three groups.

\begin{table}
    \centering
    \caption{ \label{tab:results_all_hdlss}
    Mean accuracy rates (along with standard deviations) obtained by the seven methods on all the 40 datasets. Bold values are the best mean accuracy rates obtained on each dataset.}
    \resizebox{\textwidth}{!} {
    \renewcommand{\arraystretch}{1.2}
    \begin{tabular}{|c c c c c c c c|}
        \hline
        Dataset & XGB & RF & SVM & DWD & RFSVM & COSSVM & LMNNSVM \\
        \hline\hline
        UMIST. & 0.928 $\pm$ 0.016 & 0.985 $\pm$ 0.013 & 0.365 $\pm$ 0.402 & 0.948 $\pm$ 0.009 & 0.988 $\pm$ 0.010 & 0.970 $\pm$ 0.011 & \textbf{0.989 $\pm$ 0.006} \\
        \hline
        leukemia & 0.944 $\pm$ 0.045 & 0.964 $\pm$ 0.031 & 0.950 $\pm$ 0.051 & 0.922 $\pm$ 0.057 & \textbf{0.969 $\pm$ 0.023} & 0.961 $\pm$ 0.045 & 0.956 $\pm$ 0.052 \\
        \hline
        aliz. & 0.771 $\pm$ 0.071 & 0.877 $\pm$ 0.064 & 0.890 $\pm$ 0.060 & 0.897 $\pm$ 0.061 & 0.897 $\pm$ 0.054 & \textbf{0.906 $\pm$ 0.053} & 0.903 $\pm$ 0.050 \\
        \hline
        tr45.wc & \textbf{0.970 $\pm$ 0.008} & 0.949 $\pm$ 0.012 & 0.815 $\pm$ 0.070 & 0.665 $\pm$ 0.107 & 0.954 $\pm$ 0.007 & 0.925 $\pm$ 0.008 & 0.900 $\pm$ 0.046 \\
        \hline
        laiho-2007 & 0.805 $\pm$ 0.024 & 0.805 $\pm$ 0.034 & 0.874 $\pm$ 0.059 & 0.826 $\pm$ 0.041 & 0.858 $\pm$ 0.041 & 0.879 $\pm$ 0.058 & \textbf{0.911 $\pm$ 0.041} \\
        \hline
        bittner-2000 & 0.716 $\pm$ 0.111 & 0.784 $\pm$ 0.044 & 0.753 $\pm$ 0.097 & 0.816 $\pm$ 0.035 & 0.784 $\pm$ 0.050 & \textbf{0.837 $\pm$ 0.044} & 0.800 $\pm$ 0.039 \\
        \hline
        arcene & 0.772 $\pm$ 0.026 & 0.767 $\pm$ 0.043 & 0.633 $\pm$ 0.116 & 0.820 $\pm$ 0.036 & 0.807 $\pm$ 0.038 & 0.835 $\pm$ 0.045 & \textbf{0.853 $\pm$ 0.042} \\
        \hline
        ramas. & 0.705 $\pm$ 0.037 & 0.751 $\pm$ 0.027 & 0.635 $\pm$ 0.038 & 0.669 $\pm$ 0.035 & \textbf{0.773 $\pm$ 0.021} & 0.760 $\pm$ 0.042 & 0.617 $\pm$ 0.026 \\
        \hline
        arms. & 0.942 $\pm$ 0.032 & \textbf{0.975 $\pm$ 0.015} & 0.928 $\pm$ 0.042 & 0.961 $\pm$ 0.025 & 0.972 $\pm$ 0.018 & 0.975 $\pm$ 0.015 & 0.942 $\pm$ 0.049 \\
        \hline
        su-2001 & 0.895 $\pm$ 0.026 & 0.883 $\pm$ 0.027 & 0.897 $\pm$ 0.021 & 0.883 $\pm$ 0.025 & \textbf{0.920 $\pm$ 0.024} & 0.894 $\pm$ 0.025 & 0.897 $\pm$ 0.027 \\
        \hline
        lapo. & 0.804 $\pm$ 0.044 & 0.809 $\pm$ 0.030 & 0.818 $\pm$ 0.047 & 0.815 $\pm$ 0.041 & 0.851 $\pm$ 0.039 & 0.833 $\pm$ 0.034 & \textbf{0.853 $\pm$ 0.043} \\
        \hline
        gol. & 0.939 $\pm$ 0.055 & 0.931 $\pm$ 0.031 & 0.881 $\pm$ 0.071 & 0.864 $\pm$ 0.040 & \textbf{0.944 $\pm$ 0.030} & 0.903 $\pm$ 0.031 & 0.914 $\pm$ 0.019 \\
        \hline\hline
        Dexter & 0.916 $\pm$ 0.015 & 0.925 $\pm$ 0.010 & 0.869 $\pm$ 0.030 & 0.916 $\pm$ 0.010 & 0.934 $\pm$ 0.013 & \textbf{0.936 $\pm$ 0.006} & 0.927 $\pm$ 0.013 \\
        \hline
        yeoh. & 0.842 $\pm$ 0.018 & 0.809 $\pm$ 0.015 & 0.787 $\pm$ 0.018 & 0.715 $\pm$ 0.022 & \textbf{0.848 $\pm$ 0.024} & 0.718 $\pm$ 0.023 & 0.815 $\pm$ 0.031 \\
        \hline
        tomli. & 0.715 $\pm$ 0.052 & 0.730 $\pm$ 0.050 & 0.811 $\pm$ 0.071 & 0.798 $\pm$ 0.046 & 0.763 $\pm$ 0.047 & 0.837 $\pm$ 0.049 & \textbf{0.859 $\pm$ 0.044} \\
        \hline
        khan-2001 & 0.960 $\pm$ 0.035 & \textbf{0.983 $\pm$ 0.015} & 0.955 $\pm$ 0.044 & 0.955 $\pm$ 0.034 & 0.979 $\pm$ 0.020 & 0.957 $\pm$ 0.030 & 0.974 $\pm$ 0.033 \\
        \hline
        west-2001 & 0.860 $\pm$ 0.041 & \textbf{0.884 $\pm$ 0.045} & 0.860 $\pm$ 0.048 & 0.856 $\pm$ 0.057 & 0.880 $\pm$ 0.040 & 0.860 $\pm$ 0.057 & 0.864 $\pm$ 0.060 \\
        \hline
        eating & \textbf{0.560 $\pm$ 0.022} & 0.532 $\pm$ 0.025 & 0.148 $\pm$ 0.000 & 0.551 $\pm$ 0.027 & 0.556 $\pm$ 0.018 & 0.403 $\pm$ 0.206 & 0.300 $\pm$ 0.159 \\
        \hline
        bhatt. & 0.946 $\pm$ 0.014 & 0.929 $\pm$ 0.016 & 0.937 $\pm$ 0.017 & 0.923 $\pm$ 0.022 & \textbf{0.957 $\pm$ 0.019} & 0.936 $\pm$ 0.018 & 0.934 $\pm$ 0.017 \\
        \hline
        micro. & 0.929 $\pm$ 0.032 & 0.931 $\pm$ 0.023 & 0.429 $\pm$ 0.404 & 0.912 $\pm$ 0.040 & \textbf{0.937 $\pm$ 0.024} & 0.908 $\pm$ 0.020 & 0.904 $\pm$ 0.027 \\
        \hline
        oh15.wc & 0.824 $\pm$ 0.014 & 0.825 $\pm$ 0.008 & 0.742 $\pm$ 0.041 & 0.558 $\pm$ 0.170 & \textbf{0.835 $\pm$ 0.006} & 0.803 $\pm$ 0.025 & 0.763 $\pm$ 0.010 \\
        \hline
        oh10.wc & 0.831 $\pm$ 0.012 & \textbf{0.836 $\pm$ 0.013} & 0.759 $\pm$ 0.021 & 0.654 $\pm$ 0.103 & 0.835 $\pm$ 0.017 & 0.765 $\pm$ 0.019 & 0.736 $\pm$ 0.014 \\
        \hline
        ship. & 0.841 $\pm$ 0.057 & 0.828 $\pm$ 0.051 & 0.890 $\pm$ 0.066 & 0.856 $\pm$ 0.071 & 0.882 $\pm$ 0.045 & \textbf{0.921 $\pm$ 0.044} & 0.910 $\pm$ 0.038 \\
        \hline
        cnae-9H & 0.879 $\pm$ 0.021 & \textbf{0.915 $\pm$ 0.021} & 0.888 $\pm$ 0.020 & 0.862 $\pm$ 0.034 & 0.893 $\pm$ 0.023 & 0.914 $\pm$ 0.013 & 0.818 $\pm$ 0.027 \\
        \hline
        OVA\_C. & \textbf{0.976 $\pm$ 0.003} & 0.974 $\pm$ 0.003 & 0.953 $\pm$ 0.046 & 0.968 $\pm$ 0.004 & \textbf{0.976 $\pm$ 0.003} & 0.966 $\pm$ 0.005 & 0.963 $\pm$ 0.003 \\
        \hline
        OVA\_B. & \textbf{0.972 $\pm$ 0.005} & 0.968 $\pm$ 0.006 & 0.816 $\pm$ 0.078 & 0.970 $\pm$ 0.003 & \textbf{0.972 $\pm$ 0.005} & 0.969 $\pm$ 0.004 & 0.964 $\pm$ 0.004 \\
        \hline
        imdb & 0.626 $\pm$ 0.019 & 0.701 $\pm$ 0.021 & 0.702 $\pm$ 0.025 & 0.633 $\pm$ 0.082 & 0.695 $\pm$ 0.027 & \textbf{0.710 $\pm$ 0.018} & 0.689 $\pm$ 0.025 \\
        \hline
        cnae-9 & 0.908 $\pm$ 0.012 & 0.932 $\pm$ 0.009 & 0.917 $\pm$ 0.013 & 0.861 $\pm$ 0.020 & 0.909 $\pm$ 0.009 & \textbf{0.935 $\pm$ 0.010} & 0.861 $\pm$ 0.021 \\
        \hline
        lsvt & 0.832 $\pm$ 0.038 & 0.811 $\pm$ 0.056 & 0.689 $\pm$ 0.067 & \textbf{0.871 $\pm$ 0.017} & 0.833 $\pm$ 0.048 & 0.749 $\pm$ 0.103 & 0.660 $\pm$ 0.035 \\
        \hline
        yelp & 0.709 $\pm$ 0.018 & 0.765 $\pm$ 0.017 & 0.767 $\pm$ 0.027 & 0.755 $\pm$ 0.036 & \textbf{0.776 $\pm$ 0.022} & 0.771 $\pm$ 0.024 & 0.741 $\pm$ 0.018 \\
        \hline
        amazon & 0.725 $\pm$ 0.019 & 0.789 $\pm$ 0.011 & \textbf{0.802 $\pm$ 0.011} & 0.798 $\pm$ 0.023 & 0.786 $\pm$ 0.013 & 0.795 $\pm$ 0.013 & 0.771 $\pm$ 0.012 \\
        \hline
        chow. & 0.948 $\pm$ 0.030 & 0.963 $\pm$ 0.013 & 0.963 $\pm$ 0.023 & 0.977 $\pm$ 0.021 & 0.969 $\pm$ 0.018 & \textbf{0.981 $\pm$ 0.012} & 0.915 $\pm$ 0.030 \\
        \hline\hline
        chen-2002 & 0.916 $\pm$ 0.027 & 0.937 $\pm$ 0.031 & 0.931 $\pm$ 0.030 & 0.938 $\pm$ 0.015 & \textbf{0.942 $\pm$ 0.035} & 0.928 $\pm$ 0.013 & 0.887 $\pm$ 0.031 \\
        \hline
        gina & 0.939 $\pm$ 0.005 & 0.925 $\pm$ 0.005 & 0.509 $\pm$ 0.000 & 0.852 $\pm$ 0.004 & \textbf{0.947 $\pm$ 0.004} & 0.860 $\pm$ 0.007 & 0.844 $\pm$ 0.030 \\
        \hline
        madelon & 0.760 $\pm$ 0.018 & 0.769 $\pm$ 0.017 & 0.672 $\pm$ 0.009 & 0.595 $\pm$ 0.008 & \textbf{0.799 $\pm$ 0.014} & 0.596 $\pm$ 0.015 & 0.600 $\pm$ 0.026 \\
        \hline
        scene & 0.980 $\pm$ 0.005 & 0.927 $\pm$ 0.008 & \textbf{0.990 $\pm$ 0.002} & 0.948 $\pm$ 0.042 & 0.954 $\pm$ 0.009 & 0.989 $\pm$ 0.002 & 0.973 $\pm$ 0.004 \\
        \hline
        wdbc & 0.968 $\pm$ 0.009 & 0.963 $\pm$ 0.011 & 0.937 $\pm$ 0.019 & \textbf{0.978 $\pm$ 0.004} & 0.966 $\pm$ 0.010 & 0.944 $\pm$ 0.012 & 0.956 $\pm$ 0.009 \\
        \hline
        led24 & 0.701 $\pm$ 0.005 & 0.724 $\pm$ 0.007 & 0.717 $\pm$ 0.007 & \textbf{0.729 $\pm$ 0.007} & 0.714 $\pm$ 0.006 & 0.719 $\pm$ 0.006 & 0.719 $\pm$ 0.007 \\
        \hline
        segment & 0.974 $\pm$ 0.005 & 0.970 $\pm$ 0.004 & 0.958 $\pm$ 0.007 & 0.928 $\pm$ 0.006 & \textbf{0.976 $\pm$ 0.003} & 0.956 $\pm$ 0.004 & 0.973 $\pm$ 0.005 \\
        \hline
        spam & 0.952 $\pm$ 0.004 & 0.951 $\pm$ 0.003 & 0.915 $\pm$ 0.014 & 0.884 $\pm$ 0.007 & \textbf{0.954 $\pm$ 0.005} & 0.898 $\pm$ 0.022 & 0.939 $\pm$ 0.006 \\
        \hline\hline
        \# of wins & 4 & 5 & 2 & 3 & 16 & 7 & 5 \\
        \hline
    \end{tabular}
    }
\end{table}

\subsection{Analysis of overall results}

The first observation that can be made from Table \ref{tab:results_all_hdlss} is that most of the values in bold, the best mean accuracy rates, are obtained by the methods on the right-hand side of the table, that is to say the kernel based SVM. More precisely, as shown in the last row of Table \ref{tab:results_all_hdlss}, RFSVM is the method that achieves the best performance on most of the datasets (on 16 of the 40 datasets), followed by COSSVM (on 7 of the 40 datasets), and LMNNSVM (on 5 of the 40 datasets, \textit{ex aequo} with Random Forest).

\begin{figure}[!htb]
    \centering
    \includegraphics[keepaspectratio, width=.7\columnwidth]{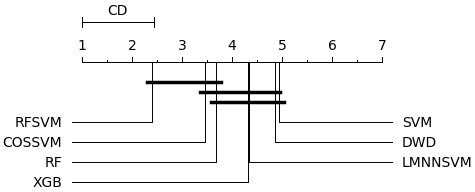}
    \caption{Critical difference diagram from the Friedman/Nemenyi test results on all the 40 datasets.}
    \label{fig:post_hoc_all}
\end{figure}

For a more precise analysis of these global results, it is necessary to look at the results of post-hoc statistical tests. Figure \ref{fig:post_hoc_all} shows the Critical Difference diagram \cite{demsar:2006} drawn from the results of the Friedman/Nemenyi test. This diagram sorts the seven methods by their average rank across the 40 datasets and indicates whether the difference between them is statistically significant: methods that are connected by a thick line are the one for which the statistical test does not reject the null-hypothesis that the classifiers perform equally well. RFSVM and COSSVM are the two best methods on average but closely followed by Random Forest. Considering statistical significance, RFSVM is the method for which the difference is statistically significant with the most methods. The differences in performance between all other methods are globally not significant according to the Friedman/Nemenyi test.

These first overall results confirm that RFSVM is particularly relevant for HDLSS classification. However, this requires further confirmation, with careful analysis of the results obtained on the HDLSS datasets, which we give in the following section. 

\subsection{Analysis of the results on HDLSS datasets}

\begin{figure}[!htb]
    \centering
    \includegraphics[keepaspectratio, width=.7\columnwidth]{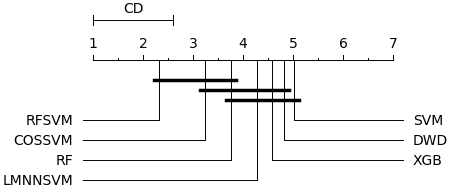}
    \caption{Critical difference diagram from the Friedman/Nemenyi test results on the HDLSS datasets ($\Omega < 1.0$).}
    \label{fig:post_hoc_hdlss}
\end{figure}

Since this work focuses on HDLSS classification, we now deepen the analysis for the HDLSS datasets, that is to say, for the datasets with $\Omega < 1.0$. Figure \ref{fig:post_hoc_hdlss} gives the Critical Difference diagram obtained by considering the HDLSS datasets only, i.e. 32 of the 40 datasets. The main difference one can observe from this diagram, compared to the one given in Figure \ref{fig:post_hoc_all}, is that LMNNSVM performs slightly better. This supports the conclusion that the kernel based approach are generally effective for HDLSS classification. It can also be noted that, in contrast, the performance of DWD is surprisingly poor compared to general-purpose machine learning methods. On the other hand, in line with the analysis given in \cite{marron:2007}, SVM is the least successful method for these tasks.

\begin{figure*}[!htb]
    \centering
    \includegraphics[keepaspectratio, width=\textwidth]{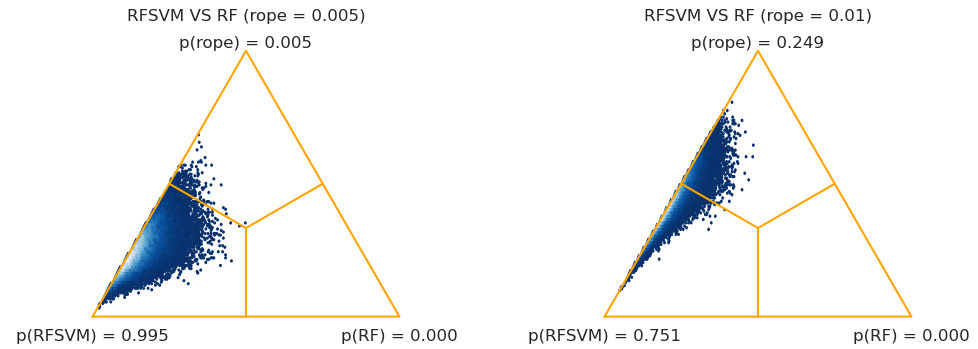} \\
    \includegraphics[keepaspectratio, width=\textwidth]{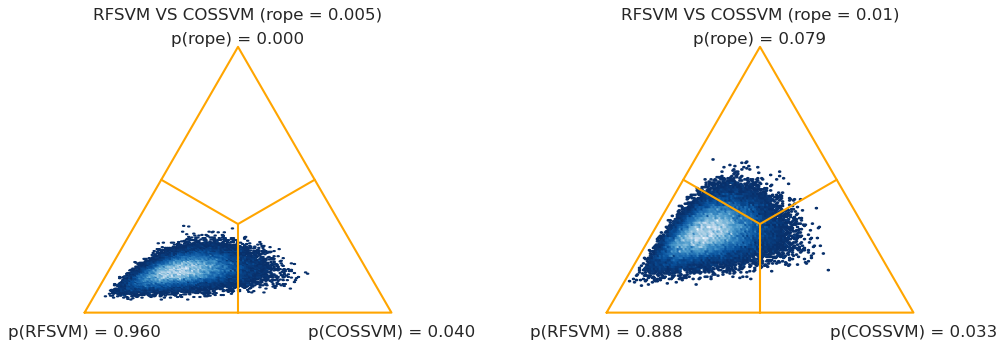} \\
    \includegraphics[keepaspectratio, width=\textwidth]{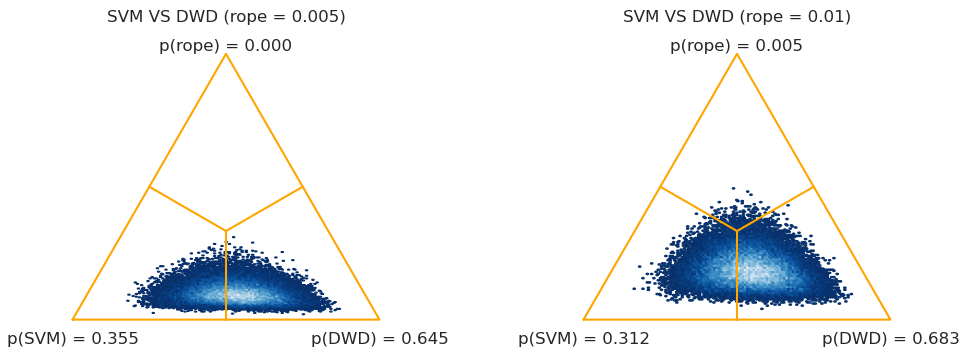}
    \caption{Examples of trinomial vectors visual representations for the Bayesian analysis on the 32 HDLSS datasets.}
    \label{fig:simplex}
\end{figure*}

The results of the Bayesian test can now be used to make a more detailed comparison, by giving the probabilities that one classifier is more accurate than another. In the following, \(p(a > b)\) denotes the probability for the classifier $a$ to be more accurate than the classifier $b$ according to the Bayesian test. Similarly, \(p(a \sim b)\) denotes the probability that classifiers $a$ and $b$ perform equally well. In a nutshell, to estimate these probabilities, the Bayesian test uses the mean differences in accuracy between classifiers $a$ and $b$ on all the datasets, and deduces a distribution for \(p(a > b)\), \(p(a \sim b)\) and \(p(b > a)\). We refer the reader to \cite{Alessio:2017} for more details on how this distribution is obtained. Then $M$ trinomial vectors of probabilities are drawn at random from this distribution. These vectors are typically represented as points in the simplex having vertices $\{(1,0,0),(0,1,0),(0,0,1)\}$. Three examples of such representations are given in Figure \ref{fig:simplex}, for three pairwise comparisons, RFSVM vs RF, RFSVM vs COSSVM and SVM vs DWD, on the HDLSS datasets. These representations allow to observe the proportion of points falling in each of the three zones corresponding to each of the three situations. By considering the number of points that fall in the three regions, one can have an estimate of all three probabilities, \(p(a > b)\), \(p(a \sim b)\) and \(p(b > a)\).

\begin{figure*}[!t]
    \centering
    \includegraphics[keepaspectratio, width=\textwidth]{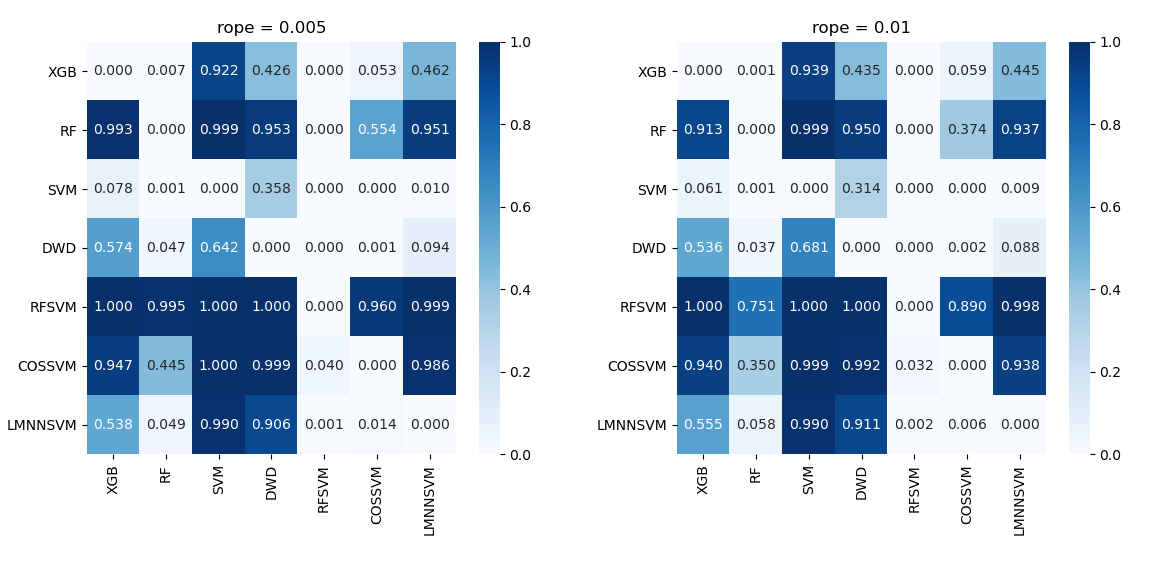}
    \caption{Pairwise Bayesian analysis for HDLSS datasets. The value in each cell is \(p(a > b)\), i.e. the probability that the row classifier $a$ outperforms the column classifier $b$.}
    \label{fig:baycomp_hdlss}
\end{figure*}

For our experimental comparison, these estimates are given as two color maps (the left one for \textit{rope}$=0.005$ and the right one for \textit{rope}$=0.01$) in Figure \ref{fig:baycomp_hdlss}. It should be read as follows: the value in the cell for the row of classifier $a$ and the column of classifier $b$ is \(p(a > b)\) according to the Bayesian test. Note that the difference between \(p(a > b)\) and \(p(b > a)\) corresponds to \(p(a \sim b)\), which is not given in these color maps. Therefore, smaller values of both \(p(a > b)\) and \(p(b > a)\) are expected with larger \textit{rope} values. However, it should be noted that this analysis is subject to random draw, and so for cases where the differences in performance between $a$ and $b$ are larger than the \textit{rope}, the opposite behavior may be observed, i.e., marginally larger probabilities occur with larger rope values.

In both color maps of Figure \ref{fig:baycomp_hdlss}, the column corresponding to the RFSVM method is the one with the lowest values. This means that overall, it is the method with the lowest probabilities to be outperformed by any other method. Thus, although RFSVM, COSSVM, and RF are not significantly different according to the Friedman/Nemenyi post-hoc test, the detailed pairwise analysis shows that both COSSVM and RF have very low probabilities of giving better results than RFSVM. Conversely, RFSVM has a high probability of being better than the cited methods. As for the comparison between SVM and DWD, these results confirm that the DWD method gives slightly better results on HDLSS datasets than the regular SVM method but with \(p(\text{DWD} > \text{SVM})\) and \(p(\text{SVM} > \text{DWD})\) being very close to $0.5$ each.

Note that for \textit{rope}$=0.01$, most of the probabilities decrease as expected, as more of the pairwise comparisons now lie into the \textit{rope}. However, the patterns observed did not have noticeable changes. 

\begin{figure}[!t]
    \centering
    \includegraphics[keepaspectratio, width=.7\columnwidth]{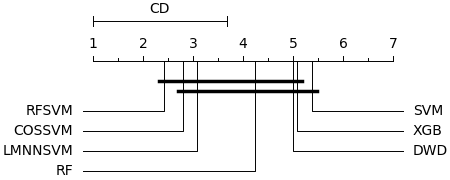}
    \caption{Critical Difference diagram from the Friedman/Nemenyi test results on the very-HDLSS datasets ($\Omega < 0.015$).}
    \label{fig:post_hoc_highly_hdlss}
\end{figure}

\begin{figure*}[!t]
    \centering
    \includegraphics[keepaspectratio, width=\textwidth]{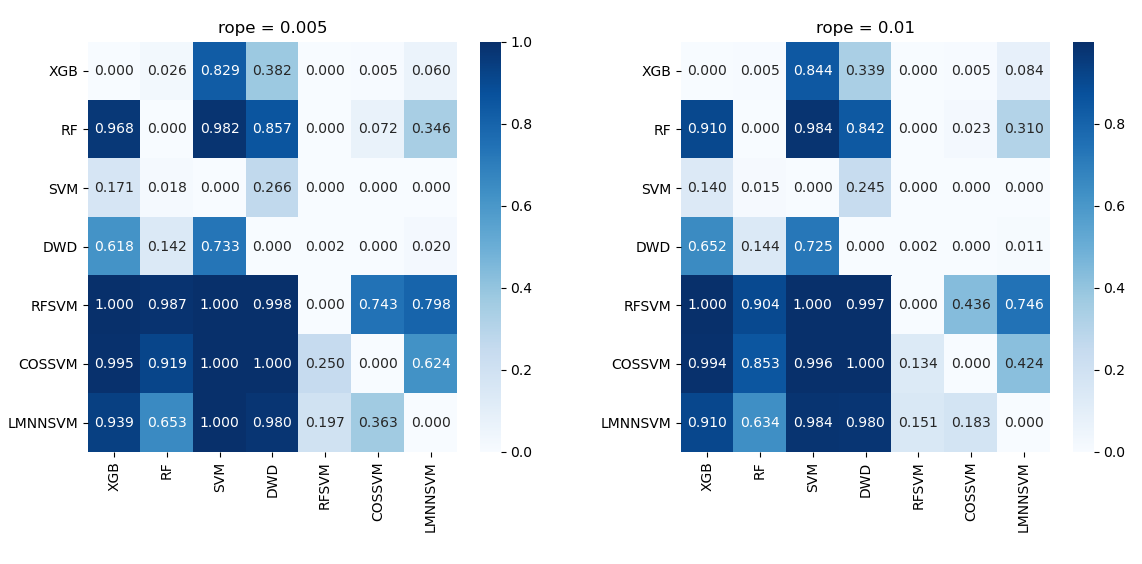}
    \caption{Pairwise Bayesian analysis for very-HDLSS datasets. The value in each cell is \(p(a > b)\), i.e. the probability that the row classifier $a$ outperforms the column classifier $b$.}
    \label{fig:baycomp_highly_hdlss}
\end{figure*}

This analysis is given considering all the datasets with $\Omega < 1.000$, that is to say, the ones at the top and in the middle part of Table \ref{tab:results_all_hdlss}. It is now interesting to focus on the 13 very-HDLSS datasets, i.e., where $\Omega < 0.015$. Figure \ref{fig:post_hoc_highly_hdlss} gives the Critical Difference diagram for these datasets only. What is interesting to note here is that LMNNSVM is now a lot more competitive and that all three similarity-based methods are clearly in the lead. The difference in average rank between these three methods and the other methods is even greater than it was in the previous results. On the other hand, these differences are considered less statistically significant by the Friedman/Nemenyi test. Nevertheless, when looking at the bayesian analysis for these 13 very-HDLSS datasets in Figure \ref{fig:baycomp_highly_hdlss}, the probability of any similarity-based methods to be outperformed by XGBoost, RF, SVM or DWD is very low (see the upper right part of the color maps). This shows that in the most extreme cases of HDLSS classification, SVM with well chosen similarity measure as kernel are particularly relevant.

\subsubsection*{A note on class imbalance}

Given the results in Table \ref{tab:datasets}, one can see that some of the datasets have quite high imbalanced ratios (high values in the `IR' column of Table \ref{tab:datasets}). In particular, three datasets show very high values in the `IR' column of Table \ref{tab:datasets}: the \textit{tr45.wc} dataset, the \textit{bhattacharjee-2001} dataset and the \textit{yeoh-2002-v2} dataset, for which the imbalance ratio is superior to 5. 
In class imbalance scenarios, it is well-known that accuracy is not a suitable performance evaluation measure. To determine whether the high IR values affect the results presented so far, we give additional results for these three specific datasets.

Table \ref{tab:f1_score} gives the results obtained by the seven methods on these three datasets in terms of F1 (top) and accuracy scores (bottom). 
F1 represents the harmonic mean between precision and recall, and it is widely applied in the assessment of imbalanced classification tasks \cite{Forman:2010}. When working with non-binary problems, we used the micro-average for F1, which computes the metric globally by counting the total number of true positives, false negatives and false positives per class.
When comparing values in both parts of the table, one can note that most of them are comparable and that the rank of all seven methods is globally similar with respect to both evaluation measures. It means that the high IR values do not call into question the conclusions drawn earlier, including for the imbalanced HDLSS datasets.

\begin{table*}[!t]
    \centering
    \caption{F1 Score (top) and Accuracy (bottom) results obtained in imbalanced datasets.}
    \resizebox{\textwidth}{!} {
    \renewcommand{\arraystretch}{1.2}
    \begin{tabular}{|c c c c c c c c|}
        \hline
        dataset & XGB & RF & SVM & DWD & RFSVM & COSSVM & LMNNSVM \\
        \hline\hline
        tr45.wc  & \textbf{0.972 $\pm$ 0.007} & 0.946 $\pm$ 0.001 & 0.745 $\pm$ 0.017 & 0.754 $\pm$ 0.017 & 0.954 $\pm$ 0.006 & 0.813 $\pm$ 0.019 & 0.784 $\pm$ 0.013 \\
        \hline
        bhattacharjee-2001 & 0.949 $\pm$ 0.020 & 0.930 $\pm$ 0.016 & 0.937 $\pm$ 0.017 & 0.932 $\pm$ 0.025 & \textbf{0.959 $\pm$ 0.021} & 0.936 $\pm$ 0.018 & 0.934 $\pm$ 0.017 \\
        \hline
        yeoh-2002-v2 & 0.848 $\pm$ 0.017 & 0.811 $\pm$ 0.017 & 0.787 $\pm$ 0.018 & 0.782 $\pm$ 0.029 & \textbf{0.849 $\pm$ 0.027} & 0.718 $\pm$ 0.023 & 0.817 $\pm$ 0.034 \\
        \hline \hline
        tr45.wc & \textbf{0.970 $\pm$ 0.008} & 0.949 $\pm$ 0.012 & 0.815 $\pm$ 0.070 & 0.665 $\pm$ 0.107 & 0.954 $\pm$ 0.007 & 0.925 $\pm$ 0.008 & 0.900 $\pm$ 0.046 \\
        \hline
        bhattacharjee-2001 & 0.946 $\pm$ 0.014 & 0.929 $\pm$ 0.016 & 0.937 $\pm$ 0.017 & 0.923 $\pm$ 0.022 & \textbf{0.957 $\pm$ 0.019} & 0.936 $\pm$ 0.018 & 0.934 $\pm$ 0.017 \\
        \hline
        yeoh-2002-v2 & 0.842 $\pm$ 0.018 & 0.809 $\pm$ 0.015 & 0.787 $\pm$ 0.018 & 0.715 $\pm$ 0.022 & \textbf{0.848 $\pm$ 0.024} & 0.718 $\pm$ 0.023 & 0.815 $\pm$ 0.031 \\
        \hline
    \end{tabular}
    }
    \label{tab:f1_score}
\end{table*}

\subsection{Analysis of the results on non-HDLSS datasets}

\begin{figure}[!h]
    \centering
    \includegraphics[keepaspectratio, width=.7\columnwidth]{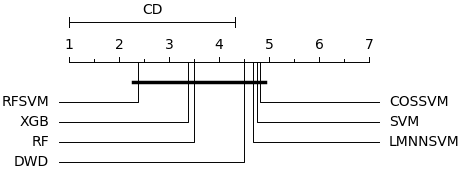}
    \caption{Critical Difference diagram from the Friedman/Nemenyi test results on the non-HDLSS datasets ($\Omega \geq 1.0$).}
    \label{fig:post_hoc_non_hdlss}
\end{figure}

Finally, we present the results on non-HLDSS datasets in order to analyze whether the similarity-based approaches are still competitive for regular classification tasks, i.e., on datasets with $\Omega \geq 1$. For that purpose, Figure \ref{fig:post_hoc_non_hdlss} gives the Critical Difference diagram on the non-HDLSS datasets, that is to say, the ones in the bottom part of Table \ref{tab:datasets}.

From this diagram, one can see that the global ranking is quite different, with in particular COSSVM and LMNNSVM being a lot less competitive. In contrast, XGBoost is not surprisingly far more accurate on average on these datasets. Nevertheless, this time, none of the differences in rank is statistically significant according to the Friedman/Nemenyi post-hoc test.

\begin{figure*}[!t]
    \centering
    \includegraphics[keepaspectratio, width=\textwidth]{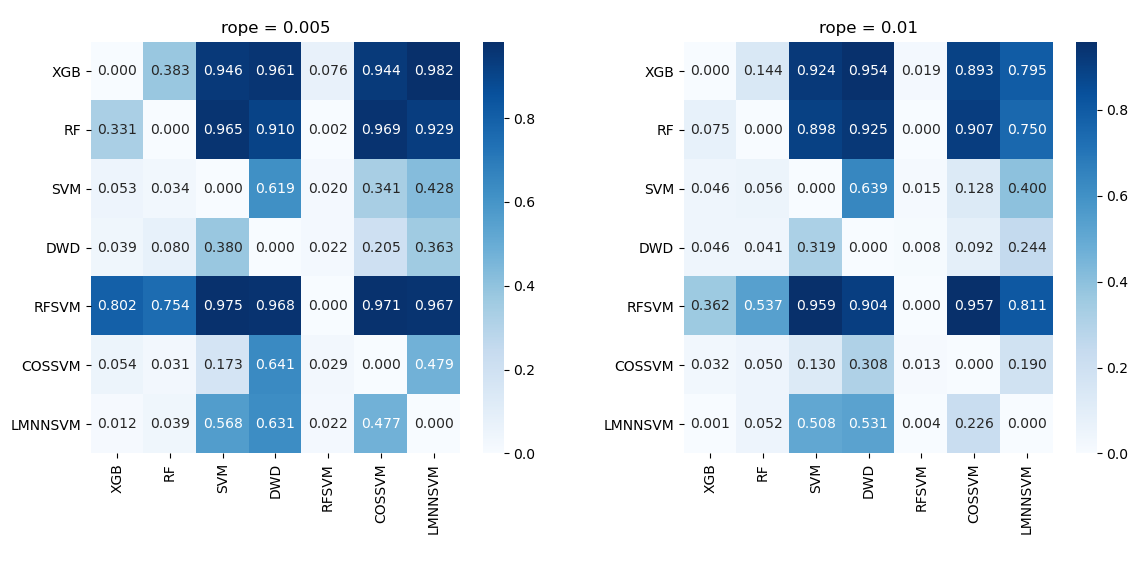}
    \caption{Pairwise bayesian analysis for non-HDLSS datasets. The value in each cell is \(p(a > b)\), i.e. the probability that the row classifier $a$ outperforms the column classifier $b$.}
    \label{fig:baycomp_non_hdlss}
\end{figure*}

However, we would like to emphasize that the RFSVM method is still ranked first on average and is particularly competitive with the state-of-the-art general-purpose classification methods, namely XGBoost and Random Forest. For a more detailed analysis, we also give the results of the Bayesian test in Figure \ref{fig:baycomp_non_hdlss}. The competitiveness of RFSVM is confirmed by the fact that the probabilities of the RFSVM column in these color maps are still very low for these datasets.

\section{Conclusion \label{sec:conclusion}}

HDLSS classification problems are unavoidable in many real-world pattern recognition problems, and having methods to provide a satisfactory solution to such problems is of crucial importance. Usually, it is faced with dimensionality reduction techniques and posterior induction of general-purpose machine learning models. However, in many situations, dimensionality reduction techniques give unsatisfactory results and genuine HDLSS learning methods are needed. 

In this work, we show that one of these methods, RFSVM, is particularly efficient, regardless of the ``degree of HDLSS'' of the problem. This method, that has been designed in our previous works for multi-view learning, is based on the use of Random Forests to estimate the similarity between the training data and on the use of these similarities as a pre-computed kernel in an SVM classifier.

To show the suitability and efficiency of RFSVM for HDLSS data, we designed a rigorous experimental comparison with different state-of-the-art classification techniques, from general-purpose techniques to state-of-the-art HDLSS methods. This comparison has been conducted on 40 different datasets, with 32 HDLSS datasets and 8 regular classification datasets for control, all publicly available. The HDLSS datasets have been carefully selected to propose a wide variety of HDLSS levels, from ``slightly HDLSS'' to ``very HDLSS''. Two statistical analyses of the results have been conducted to support the superiority of RFSVM over the others, a frequentist analysis with the Friedman statistical test along with the Nemenyi post-hoc test, and a bayesian analysis with a Bayesian sign test. Both statistical analyses confirm the superiority of RFSVM over its competitors, but with the interesting nuance that, more globally, similarity-based approaches are the most robust to extremely HDLSS learning conditions.

Based on the success of these similarity-based classification methods, we believe it is worthwhile to apply them to more specific tasks where HDLSS datasets are known to be challenging, as anomaly/outlier detection, novelty detection, or data domain description. Detecting outliers, for example, is of particular importance in HDLSS problems since without much data for training, outliers may have an important influence on the result. Estimating distances between points, which is crucial for detecting outliers, is however difficult in the HDLSS context. We believe that the Random Forest Kernel approach could be an efficient alternative in this context.

Finally, many naturally HDLSS real-world applications have a characteristic that is usually a challenge in machine learning: data sparsity. 
\rev{While Random Forests are known to be robust to high dimensions, they are also known to suffer from data sparsity. However, these methods are frequently used on sparse HDLSS data as they embed efficient pre-processing and/or interpretability tools. For example, they may serve as feature selectors in a sparse data classification context, as in \cite{Diaz:2006} where they are used for gene selection and classification of microarray data. For this reason, we believe it is relevant to further investigate the behavior of our method in the context of sparse data learning.}

\section{Acknowledgments}
This work is part of the DAISI project, co-financed by the European Union with the European Regional Development Fund (ERDF) and by the Normandy Region.

\bibliographystyle{elsarticle-num}
\bibliography{hdlss_preprint}

\end{document}